\documentclass[runningheads]{llncs}

\usepackage[utf8]{inputenc} % allow utf-8 input
\usepackage[T1]{fontenc}    % use 8-bit T1 fonts
\usepackage{hyperref}       % hyperlinks
\usepackage{url}            % simple URL typesetting
\usepackage{booktabs}       % professional-quality tables
\usepackage{amsfonts}       % blackboard math symbols
\usepackage{nicefrac}       % compact symbols for 1/2, etc.
\usepackage{microtype}  
\usepackage{graphicx}
\usepackage{multirow}
\usepackage{makecell}
\usepackage{tabularx}
\usepackage{textcomp}
\usepackage{enumitem}
\usepackage[table]{xcolor}
\usepackage{float}
\usepackage{adjustbox}
\usepackage[htt]{hyphenat}
\usepackage[square,numbers]{natbib}
\begin{document}
\title{Long-Context Encoder Models for Polish Language Understanding}
\titlerunning{Long-Context Encoder Models for Polish Language Understanding}

\author{Sławomir Dadas\inst{1} \and Rafał Poświata\inst{1} \and Marek Kozłowski\inst{1} \and Małgorzata Grębowiec\inst{1} \and Michał Perełkiewicz\inst{1} \and Paweł Klimiuk\inst{2} \and Przemysław Boruta\inst{2}}
\authorrunning{S. Dadas et al}
\institute{
National Information Processing Institute, Warsaw, Poland\\
\and
PKO Bank Polski, Warsaw, Poland\\
\email{\{sdadas,rposwiata,mkozlowski,mgrebowiec,mperelkiewicz\}@opi.org.pl}\\
\email{\{pawel.klimiuk,przemyslaw.boruta\}@pkobp.pl}
}

\maketitle 
\begin{abstract}
While decoder-only Large Language Models (LLMs) have recently dominated the NLP landscape, encoder-only architectures remain a cost-effective and parameter-efficient standard for discriminative tasks. However, classic encoders like BERT are limited by a short context window, which is insufficient for processing long documents. In this paper, we address this limitation for the Polish by introducing a high-quality Polish model capable of processing sequences of up to 8192 tokens. The model was developed by employing a two-stage training procedure that involves positional embedding adaptation and full parameter continuous pre-training. Furthermore, we propose compressed model variants trained via knowledge distillation.  The models were evaluated on 25 tasks, including the KLEJ benchmark, a newly introduced financial task suite (FinBench), and other classification and regression tasks, specifically those requiring long-document understanding. The results demonstrate that our model achieves the best average performance among Polish and multilingual models, significantly outperforming competitive solutions in long-context tasks while maintaining comparable quality on short texts.

\end{abstract}

\section{Introduction}
Since the inception of the Transformer architecture \citep{vaswani2017attention}, it has predominated the landscape of neural language modeling. Language models utilizing this architecture fall into three categories: encoder-only (such as BERT \citep{devlin2019bert} or RoBERTa \citep{liu2019roberta}), encoder-decoder (such as T5 \citep{raffel2020exploring}), and decoder-only (most modern LLMs \citep{raiaan2024review}). In the years immediately following BERT's release, the research community focused predominantly on encoders, which yielded state-of-the-art performance on numerous discriminative tasks, including text classification or named entity recognition (NER). With the growth in computational power and the scaling of models to larger sizes, attention gradually shifted towards decoders. Due to their generative capabilities, these models, given a sufficient number of parameters, can generalize to solve a broader range of tasks than encoders \citep{brown2020language}. Nevertheless, encoders have not been rendered obsolete. They remain more cost-effective to deploy and parameter-efficient compared to decoders, delivering comparable performance on discriminative tasks at a fraction of the size of LLMs \citep{kocon2023chatgpt}.

One of the important issues limiting the utility of older encoders was their restricted context window, which for the original BERT and RoBERTa models amounted to only 512 tokens, corresponding to approximately one page of text. This capacity is insufficient for many modern applications that require the processing of longer documents. In response to these limitations, several new encoder-only models have recently emerged, primarily for English, offering longer context windows and a set of architectural improvements. The most prominent among these are ModernBERT \citep{warner2025smarter}, NeoBERT \citep{breton2025neobert}, and MosaicBERT \citep{portes2023mosaicbert}. Regarding multilingual models, releases include mmBERT \citep{marone2025mmbertmodernmultilingualencoder}, based on the same architecture as ModernBERT, and EuroBERT \citep{boizard2025eurobert}.

In this paper, we present a new encoder-only model for the Polish language with an extended context, capable of processing texts up to 8192 tokens in length. Our approach involved leveraging an existing high-quality Polish RoBERTa encoder and subsequently adapting it to support long documents. The encoder was trained in two stages: first by extending and adapting the positional embeddings, and then by fine-tuning the entire model. We enhanced the Polish RoBERTa architecture implementation to support Flash Attention \citep{dao2022flashattention} and contamination-free packing, which increased training efficiency and enabled context extension while maintaining performance on short tasks. We also conducted a comprehensive evaluation across 25 downstream tasks, comparing our model against both older and modern encoders supporting the Polish language.

\textbf{Our contributions are as follows}: \textbf{1)} We trained and released a Polish encoder-only model with an extended context of up to 8192 tokens. \textbf{2)} Employing knowledge distillation techniques, we created compressed versions of the model with layer counts reduced by 50\% and 75\%, designed for efficiency-critical applications such as edge devices. \textbf{3)} We conducted an evaluation of our models, comparing them against other publicly available encoder-only models supporting the Polish language. The evaluation encompassed 25 tasks, some of which were prepared specifically for this study. Notably, we introduced FinBench, a suite of 7 Polish-language tasks from the banking and finance domain.

\section{Related work}
The development of neural language models for Polish has not been as intensive as for high-resource languages such as English or Chinese. The most significant activity in this area could be observed during the early years of Transformer development, shortly after the release of the English version of BERT \citep{devlin2019bert}. It was then that Polish-language models emerged, trained on corpora ranging from tens to hundreds of gigabytes of text, which were considered large by the standards of the time. This primarily refers to two versions of the HerBERT model \citep{rybak2020klej,mroczkowski-etal-2021-herbert} and two versions of Polish RoBERTa \citep{dadas2020pre}. In addition to encoder-only models, encoder-decoder models were also released, notably plT5 \citep{chrabrowa2022evaluation} and Polish BART \citep{polish-nlp-resources}, as well as small Polish GPT-2 decoders in three sizes \citep{polish-nlp-resources}. In subsequent years, few general-purpose models were developed. Research focused primarily on fine-tuning existing models to solve practical problems or creating domain-specific language models such as TrelBERT \citep{szmyd-etal-2023-trelbert}, trained on a Polish Twitter corpus.

A second wave of interest in Polish models arrived with the popularization of Large Language Models (LLMs). This surge in interest is not unique to Poland but extends to many other countries. However, it focuses almost exclusively on decoder-only models, driven by the investments and high expectations associated with the development of this technology. In the case of Poland, two initiatives stand out in terms of scale and resources: the PLLuM \citep{kocon2025pllum} and Bielik \citep{ociepa2025bielik11bv2technical} model families, although many more modest attempts to train Polish LLMs have also been undertaken. So far, the community and scientific interest in LLMs has not transitioned to encoder-only and encoder-decoder architectures.

\section{Training}

\subsection{Model with extended context}
Our solution is based on the existing \emph{polish-roberta-large-v2} model \citep{dadas2020pre}, chosen for its state-of-the-art performance on the KLEJ benchmark \citep{rybak2020klej}. In the first step, we modified the model’s architecture by extending the positional embedding layer to 8192 positions. Since the new embeddings were initialized with random values, further training was required to enable the model to handle longer documents. Prior to training, we introduced two architectural modifications to optimize the learning process for longer texts. The first modification involved implementing the support for Flash Attention \citep{dao2022flashattention}, which reduced both memory consumption and computational overhead. The second modification was the implementation of contamination-free packing. Packing is a common technique that involves concatenating multiple shorter documents until the model's maximum context size is reached. In the naive approach, documents are simply concatenated and separated by a special token. However, this method allows content from unrelated documents to cross-contaminate via the attention mechanism, which can negatively affect the quality of the trained model. In our implementation, we constrained the attention mechanism to prevent it from crossing document boundaries within the same sequence. Keep in mind that the introduced modifications are primarily intended to reduce memory consumption and increase training efficiency, while remaining numerically consistent with the original architecture. Therefore, it is possible to use the trained model with the standard implementation of the RoBERTa model.

The model was trained on a Polish language corpus comprising 150 billion tokens. The training process lasted for five epochs and was divided into two stages: the first stage comprised one epoch, while the second covered the remaining four. We applied the same hyperparameters in both stages. The model was trained using the AdamW optimizer with a polynomial learning rate scheduler, a maximum learning rate of 2e-5, and a warmup phase of 500 batches. The total batch size was 128 with a sequence length of 8192, which resulted in a batch size exceeding 1 million tokens. We utilized masked language modeling (MLM) with a masking rate of 20\% - which is 5\% higher than that used for BERT \citep{devlin2019bert}, RoBERTa \citep{liu2019roberta}, and Polish RoBERTa models \citep{dadas2020pre}, and identical to the rate used in the newer NeoBERT \citep{breton2025neobert} architecture. We also employed whole word masking, where any continuous alphanumeric string was treated as a word. The training stages proceeded as follows:

\begin{table}[h]
  \centering
  \caption{Comparison of encoder architectures for the Polish language. The upper section of the table presents our trained models with extended context, while the lower section lists popular base-sized models supporting Polish. We report the model size in terms of total parameters and parameters excluding the embedding layer, the number of encoder layers, and the estimated number of floating-point operations required for a single forward pass with a sequence length of 512 tokens.}
  \aboverulesep=0ex
  \belowrulesep=0ex
  \setlength{\tabcolsep}{5pt}
  \renewcommand{\arraystretch}{0.8}
  \begin{tabular}{|l|c|c|c|c|c|}
    \toprule
    \textbf{Model} & \textbf{Size} & \makecell{\textbf{Size excluding} \\ \textbf{embedding}} & \makecell{\textbf{Encoder} \\ \textbf{layers}} & \textbf{GFLOPs} \\
    \toprule
    polish-roberta-8k & 443M & 303M & 24 & 167.5 \\
    polish-roberta-8k-12L & 291M & 152M & 12 & 83.8 \\
    polish-roberta-8k-6L & 216M & 77M & 6 & 41.9 \\
    \hline
    sdadas/polish-roberta-base-v2 & 124M & 86M & 12 & 48.3 \\
    allegro/herbert-base-cased & 124M & 86M & 12 & 48.3 \\
    FacebookAI/xlm-roberta-base & 278M & 86M & 12 & 48.3 \\
    \bottomrule
  \end{tabular}
  \label{tab:distillation}
\end{table}

\begin{enumerate}[wide,labelwidth=0pt,labelindent=0pt,itemsep=0pt,topsep=5pt]
\item{In the first stage, all model weights were frozen, and only the new positional embedding layer was trained. This preliminary stage proved necessary because training all weights immediately after extending the context caused excessive gradient fluctuations, which detrimentally affected the previously trained weights of the base model. The objective of this stage was to adapt the new positional embeddings to align with the existing model.}
\item{In the second stage, the model was trained on the same corpus for four epochs. In this stage, all model parameters were updated.}
\end{enumerate}

Since the model's ability to handle short versus long predictive tasks changed during training, the final model is the result of merging selected checkpoints from three epochs of the second training stage (specifically from epochs 2, 3, and 4). This ensured balanced prediction quality for both short and long texts.

\subsection{Knowledge distillation}
In certain practical encoder applications, there is a need to deploy compact models capable of operating on edge devices with constrained computational and memory resources. Consequently, our experiments also focused on compressing the newly trained \emph{polish-roberta-8k} model into smaller variants suitable for such scenarios. These experiments involved downsizing the model by removing a subset of encoder layers, followed by utilizing knowledge distillation during the subsequent training of these compressed versions to recover the performance of the original model. We developed models in which the layer count was reduced by 50\% (from 24 to 12) and by 75\% (to 6), respectively. A summary of these models, along with a comparison against other compact models supporting the Polish language, is presented in Table \ref{tab:distillation}. It can be observed that for smaller models, the number of parameters in the embedding layer begins to dominate the remaining model parameters; however, this layer has a negligible impact on the computational cost of the forward pass. Approximately, \emph{polish-roberta-8k-12L} achieves 57\% of the throughput of base-sized models, whereas \emph{polish-roberta-8k-6L} achieves 115\%.

For the training of the smaller model versions, we opted for a simple knowledge distillation approach utilizing mean squared error (MSE) between the token representations in the final layer of the original model (teacher) and its compressed version (student). While higher-quality students could theoretically be obtained using attention matrix distillation, this approach is infeasible for sequence lengths of 8192 tokens due to the quadratic growth in computational and memory complexity - such encoder distillation has been performed only for short contexts of 128 \citep{wang2023distill} or 256 \citep{wang2020minilm} tokens.

Distillation was performed for one epoch on the same corpus on which the original model was trained. The hyperparameters were also similar, we only reduced the batch size to 64 sequences, and the maximum learning rate was increased to 5e-5. Additionally, the smallest model with 6 encoder layers was prepared in two versions - a simple one and a progressive one. In the simple version, we performed a direct distillation from the \emph{polish-roberta-8k} model. In the progressive version, we first performed distillation to a 12-layer model, and only then did we reduce that model to 6 layers and perform another iteration of distillation.

\section{Evaluation}
In this section, we present the evaluation results of our models with extended context and compare them with other models supporting the Polish language. The evaluation encompasses 25 tasks, including regression, as well as single-label and multi-label classification. Given that our models are intended for deployment within the banking sector, in addition to verifying quality on publicly available datasets, we have also prepared a specialized benchmark focusing exclusively on the financial and banking domain. The first part of the section describes the datasets used in the evaluation, the second part presents detailed results of the \emph{polish-roberta-8k} model compared to leading Polish encoders, next we provide a broader comparison of available Polish and multilingual models, and finally, the results of bank's internal evaluation is presented.

\subsection{Evaluation data}
The tasks included in the evaluation can be categorized into three groups. The first consists of one of the most well-known Polish benchmarks for encoder models: \emph{KLEJ}. The second group comprises banking and finance datasets prepared by us during our project work and compiled into a benchmark named \emph{FinBench}. The third group consists of other Polish-language tasks that are not part of a larger benchmark. The characteristics of all included datasets are presented below.

\noindent \textbf{KLEJ Benchmark} - KLEJ \citep{rybak2020klej} is a popular benchmark consisting of nine datasets. It includes sentiment analysis tasks (PolEmo-IN, PolEmo-OUT, Allegro Reviews), named entity recognition (NKJP-NER), harmful tweet classification (CBD), and four tasks involving the determination of various types of semantic relations between text pairs (CDSC-R, CDSC-R, DYK, PSC).

\noindent \textbf{Financial benchmark (FinBench)} - Given that the model was trained for its deployment in a bank, it was necessary to evaluate its ability to solve tasks within the financial and banking domain. Therefore, as part of our work, we created new Polish-language tasks focusing exclusively on this domain. The tasks were either created from scratch using Polish texts or translated from existing English datasets. For datasets originally in English, the DeepSeek-V3 \citep{liu2024deepseek} model was used for translation. The following tasks were prepared as part of the benchmark:
\begin{itemize}[wide,labelwidth=0pt,labelindent=0pt,itemsep=0pt,topsep=5pt]
\item[$\bullet$]{\textbf{Banking-Short} and \textbf{Banking-Long} - Two classification tasks involving the assignment of a text from the banking and financial domain to one of 14 thematic categories. The texts and their categories originate from the \emph{bankier.pl} web service. The datasets consist of 11,200 texts divided into training (7,000), validation (1,400), and test (2,800) sets. The difference between the tasks lies in the text length. In the \emph{banking-short} version, classification is performed based solely on titles consisting of a dozen to several dozen words, whereas the \emph{banking-long} version utilizes full article contents, which can exceed the size of several thousand tokens.}
\item[$\bullet$]{\textbf{Banking77} - A customer intent classification task composed of questions regarding banking services. Each query can be classified into one of 77 intents. The dataset consists of over 13,000 records and was machine-translated from the English version \citep{Casanueva2020}.}
\item[$\bullet$]{\textbf{FiQA} - The task involves assessing information regarding a publicly traded stock from a short press note. The scores are continuous and can take both negative values (information with a negative significance for the company) and positive values (positive news for the company). Model quality is measured using the R\textsuperscript{2} metric. The dataset consists of over a thousand texts, 234 of which constitute the test set, and was translated from the English version \citep{maia201818}.}
\item[$\bullet$]{\textbf{FPB} - The task involves assigning a general sentiment (negative, neutral, positive) to a press note concerning markets or specific companies. The dataset consists of approximately 4,846 records and was translated into Polish from the English version \citep{malo2014good}.}
\item[$\bullet$]{\textbf{GCN} - The dataset consists exclusively of information regarding the gold market, and each record is described by a set of nine attributes defining the topics discussed in a given piece of information (e.g., whether the information concerns future prediction or retrospective analysis, whether asset prices are compared, or whether the text discusses a price increase). This task is an example of multi-label classification. To measure model quality, we use the weighted F1 score, where the weight is proportional to the frequency of a given feature in the dataset. The dataset totals over 11,000 records. The data was translated from the English version \citep{sinha2021impact}.}
\item[$\bullet$]{\textbf{Stooq} - The dataset consists of analyses regarding the market situation sourced from the \emph{stooq.pl} web page. The task involves determining the general sentiment of a commentary as positive, negative, or neutral. The total number of records in the dataset is 1,813.}
\end{itemize}

\noindent \textbf{Other tasks} - In addition to the benchmarks described above, we also used other predictive tasks available for the Polish language for evaluation. These include three tasks concerning semantic relations between texts: SICK-E \citep{dadas2020evaluation} (textual entailment), SICK-R \citep{dadas2020evaluation} (semantic similarity), and PPC \citep{dadas2022training} (paraphrase detection). Furthermore, we included datasets covering thematic classification: 8TAGS \citep{dadas2020evaluation} (eight thematic categories of headlines from the wykop.pl service) and EURLEX \citep{chalkidis-etal-2019-large} (multi-label classification of European Union legal acts). Two datasets relate to the emotion understanding: TwitterEMO \citep{bogdanowicz2023twitteremo} (multi-label emotion classification in tweets) and IMDB \citep{maas2011learning} (sentiment analysis of movie reviews, we translated part of the original dataset from English). The evaluation set is complemented by the MIPD \citep{modzelewski2024mipd} (detection of manipulation and disinformation in Polish press articles) and BAN-PL \citep{kolos2024ban} (content moderation classification for the wykop.pl service) tasks.

\subsection{Per-task evaluation}
In the first experiment, we compared the performance of our model trained with an extended context against two of the most popular Polish encoders: \emph{herbert-large-cased} \citep{mroczkowski-etal-2021-herbert} and \emph{polish-roberta-large-v2} \citep{dadas2020pre}. The context length of these models is 512 tokens, which is significantly shorter than the 8192 tokens supported by our model. However, it should be noted that the majority of publicly available datasets used for evaluating encoders consist of short texts that fit within the 512-token limit. Moreover, for tasks such as classification or regression, the model does not necessarily need to process the entire text to achieve high prediction quality. According to our analysis, out of the 25 datasets included in the experiments, only four consist of texts that significantly exceed the length of 512 tokens. Three of these are \emph{Banking-Long}, \emph{MIPD}, and \emph{EURLEX}. The fourth, \emph{IMDB}, is a subset of an English dataset of the same name that we translated into Polish using DeepSeek-V3 \citep{liu2024deepseek}. To simulate a sentiment analysis task with a longer context, we selected only those reviews from this dataset exceeding 2,000 characters in length, discarding shorter texts.

For each task and model, we performed five independent fine-tuning runs with randomly selected seeds, and the reported results represent the average values. For tasks from the KLEJ benchmark, we used hyperparameters from the original fine-tuning scripts for \emph{polish-roberta-large-v2}. For the remaining tasks, we applied the same hyperparameters: 10 epochs, a batch size of 32, a scheduler with a warmup phase covering 6\% of the total iterations, a maximum learning rate of 1e-5, and polynomial decay.

\begin{table}[h]
  \centering
  \caption{A detailed comparison of fine-tuning \emph{polish-roberta-8k}, \emph{polish-roberta-large-v2}, and \emph{herbert-large-cased} models on 25 tasks. The reported scores are averaged across five independent training runs.}
  \aboverulesep=0ex
  \belowrulesep=0ex
  \setlength{\tabcolsep}{2pt}
  \renewcommand{\arraystretch}{0.7}
  \begin{tabular}{|l|l|l|l|ccc|}
    \toprule
    \textbf{Task type} & \textbf{Domain} & \textbf{Metric} & \textbf{Task name} & \scalebox{.85}[1.0]{\textbf{HerBERT}} & \scalebox{.85}[1.0]{\makecell{\textbf{RoBERTa} \\ \textbf{V2}}} & \scalebox{.85}[1.0]{\makecell{\textbf{RoBERTa} \\ \textbf{8K}}}  \\
    \toprule
    \multicolumn{7}{|l|}{\textbf{KLEJ Benchmark}} \\
    \hline
    single-label & mixed & accuracy & \cellcolor{cyan!10} NKJP-NER & \color{blue} 96.07 & 95.75 & 95.64 \\
    single-label & semantics & accuracy & \cellcolor{cyan!10} CDSC-E & \color{blue} 94.78 & 94.16 & 94.28 \\
    regression & semantics & spearman & \cellcolor{cyan!10} CDSC-R & 95.01 & 95.25 & \color{blue} 95.33 \\
    single-label & soc.media & binary-F1 & \cellcolor{cyan!10} CBD & 70.21 & 73.10 & \color{blue} 73.23 \\
    single-label & reviews & accuracy & \cellcolor{cyan!10} POLEMO-IN & 91.39 & \color{blue} 93.55 & 93.05 \\
    single-label & reviews & accuracy & \cellcolor{cyan!10} POLEMO-OUT & 81.66	& \color{blue} 83.81 & 83.64 \\
    single-label & mixed & binary-F1 & \cellcolor{cyan!10} DYK & 73.31 & \color{blue} 74.87 & 74.05 \\
    single-label & news & binary-F1 & \cellcolor{cyan!10} PSC & \color{blue} 98.85 & 98.37 & 98.56 \\
    regression & reviews & 1-wMAE & \cellcolor{cyan!10} AR & 89.23 & \color{blue} 89.36 & 88.91 \\
    \hline
    \multicolumn{7}{|l|}{\textbf{FinBench}} \\
    \hline
    single-label & finance & accuracy & \cellcolor{magenta!10} Banking-Short & 81.80 & 81.69 & \color{blue} 81.99 \\
    single-label & finance & accuracy & \cellcolor{magenta!10} Banking-Long & 86.64 & 87.89 & \color{blue} 88.35 \\
    single-label & finance & accuracy & \cellcolor{magenta!10} Banking77 & \color{blue} 92.76 & 92.45 & 92.74 \\
    regression & finance & r2-score & \cellcolor{magenta!10} FiQA & 61.20 & 65.71 & \color{blue} 68.43 \\
    single-label & finance & accuracy & \cellcolor{magenta!10} FPB & 84.99 & 85.26 & \color{blue} 85.42 \\
    multi-label	& finance & weighted-F1 & \cellcolor{magenta!10} GCN & \color{blue} 95.25 & 95.04 & 94.97 \\
    single-label & finance & accuracy & \cellcolor{magenta!10} Stooq & 82.53 & \color{blue} 85.07 & 84.41 \\
    \hline
    \multicolumn{7}{|l|}{\textbf{Other tasks}} \\
    \hline
    single-label & soc.media & accuracy & \cellcolor{green!10} 8TAGS & 81.16 & \color{blue} 81.64 & 81.44 \\
    single-label & soc.media & accuracy & \cellcolor{green!10} BAN-PL & 93.25 & 93.80 & \color{blue} 93.99 \\
    multi-label	& news & weighted-F1 & \cellcolor{green!10} MIPD & 66.79 & 67.27 & \color{blue} 68.50 \\
    single-label & semantics & accuracy & \cellcolor{green!10} PPC & 89.78 & \color{blue} 89.96 & 89.48 \\
    single-label & semantics & accuracy & \cellcolor{green!10} SICK-E & 87.33 & 88.33 & \color{blue} 88.96 \\
    regression & semantics & spearman & \cellcolor{green!10} SICK-R & 84.37	& 85.93 & \color{blue} 86.54 \\
    multi-label	& soc.media	& weighted-F1 & \cellcolor{green!10} TwitterEMO & 70.51	& \color{blue} 70.70 & 70.60 \\
    single-label & reviews & accuracy & \cellcolor{green!10} IMDB & 93.55 & 94.36 & \color{blue} 96.03 \\
    multi-label	& law & weighted-F1 & \cellcolor{green!10} EURLEX & 79.68 & 79.19 & \color{blue} 79.77 \\
    \hline
    \multicolumn{4}{|l|}{\textbf{Average}} & 84.88 & 85.70 & \color{blue} 85.93 \\
    \bottomrule
  \end{tabular}
  \label{tab:tasks}
\end{table}

Detailed evaluation results for the three models, broken down by individual tasks, are presented in Table \ref{tab:tasks}. The table also includes information additional such as the task type, text domain, and the metric used for evaluation. All metrics range from 0 to 100\%, with the exception of R\textsuperscript{2} and Spearman's rank correlation, which can take negative values (though this did not occur in our experiments). The table is divided into sections corresponding to the three task groups described earlier. The best result for each task is highlighted in blue. The \emph{polish-roberta-8k} model achieves the highest average score and outperforms the others on 12 out of 25 tasks, although the difference compared to \emph{polish-roberta-large-v2} is not substantial. Unsurprisingly, we observe significant differences in favor of the new model on the four datasets with long texts mentioned earlier. On short texts, the models yield comparable results in most cases. The \emph{herbert-large-cased} model achieves a slightly lower average score than the other two models, lagging behind them particularly on tasks outside the KLEJ benchmark.

\subsection{Polish and multilingual models}

In the second stage of our experiments, we conducted a broader evaluation which, in addition to the three previously tested models, also included distilled versions of \emph{polish-roberta-8k}, as well as publicly available Polish and multilingual encoders of various sizes. The results of this evaluation are presented in Table \ref{tab:models}. Additional models included in this experiment comprise the well-known multilingual \emph{XLM-Roberta} \citep{conneau2020unsupervised}, as well as modern architectures such as \emph{mmBERT} \citep{marone2025mmbertmodernmultilingualencoder} and \emph{EuroBERT} \citep{boizard2025eurobert}. Given that the literature suggests training these new architectures with slightly higher learning rates, we performed two fine-tuning runs for these models: one with the same learning rate as the other models, and a second with a learning rate of 3e-5. We reported the better of the obtained results in the table. The evaluated models were divided into two subgroups based on their size, with a cutoff threshold of 300 million parameters. The table reports average results for all tasks, as well as breakdowns into specific task subgroups: KLEJ and FinBench, other tasks, and tasks with long texts. The best result in each group is marked in blue, and the second-best in red.

\begin{table}[h]
  \centering
  \caption{Comparison of Polish and multilingual models.}
  \aboverulesep=0ex
  \belowrulesep=0ex
  \setlength{\tabcolsep}{2pt}
  \renewcommand{\arraystretch}{0.75}
  \begin{tabular}{|l|c|ccc|c|c|}
    \toprule
    \textbf{Model} & \textbf{Size} & \textbf{KLEJ} & \textbf{FinBench} & \makecell{\textbf{Other} \\ \textbf{tasks}} & \makecell{\textbf{Long} \\ \textbf{tasks}} & \makecell{\textbf{All} \\ \textbf{tasks}} \\
    & & 9 tasks & 7 tasks & 9 tasks & 4 tasks & 25 tasks \\
    \toprule
    \multicolumn{7}{|l|}{\textbf{Bigger models (>300M params)}} \\
    \hline
    EuroBERT/EuroBERT-610m & 608M & 80.10 & 78.36 & 78.10 & \color{blue} 83.24 & 78.89 \\
    jhu-clsp/mmBERT-base & 307M & 83.17 & 76.59 & 80.14 & 82.08 & 80.24 \\
    FacebookAI/xlm-roberta-large & 560M & 87.29 & 79.17 & 82.24 & 81.10 & 83.20 \\
    allegro/herbert-large-cased & 355M & 87.83 & 83.60 & 82.94 & 81.67 & 84.88 \\
    sdadas/polish-roberta-large-v2 & 435M & \color{blue} 88.69 & \color{purple} 84.73 & \color{purple} 83.46 & 82.18 & \color{purple} 85.70 \\
    \rowcolor[HTML]{f0f0f0} polish-roberta-8k & 443M & \color{purple} 88.52 & \color{blue} 85.19 & \color{blue} 83.92 & \color{purple} 83.16 & \color{blue} 85.93 \\
    \hline
    \multicolumn{7}{|l|}{\textbf{Smaller models (<300M params)}} \\
    \hline
    EuroBERT/EuroBERT-210m & 212M & 77.16 & 76.68 & 72.48 & 80.42 & 75.34 \\
    FacebookAI/xlm-roberta-base & 278M & 84.46 & 76.63 & 76.29 & 74.42 & 79.33 \\
    allegro/herbert-base-cased & 124M & 85.83 & 79.20 & 78.59 & 77.08 & 81.37 \\
    \rowcolor[HTML]{f0f0f0} polish-roberta-8k-6L (simple) & 216M & 84.55 & 81.53 & 80.15 & \color{purple} 81.16 & 82.12 \\
    sdadas/polish-roberta-base-v2 & 124M & \color{purple} 86.75 & 81.07 & 79.69 & 77.30 & 82.62 \\
    \rowcolor[HTML]{f0f0f0} polish-roberta-8k-6L (progressive) & 216M & 85.20 & \color{purple} 82.43 & \color{purple} 80.57 & 80.95 & \color{purple} 82.76 \\
    \rowcolor[HTML]{f0f0f0} polish-roberta-8k-12L & 291M & \color{blue} 87.21 & \color{blue} 83.73 & \color{blue} 82.61 & \color{blue} 81.86 & \color{blue} 84.58 \\
    \hline
    \bottomrule
  \end{tabular}
  \label{tab:models}
\end{table}

In the large model group, \emph{polish-roberta-8k} maintains the lead. New models released within the last year perform poorly with the Polish language. \emph{EuroBERT-610m} and \emph{mmBERT-base} occupy the last two places, significantly lagging behind models trained exclusively on Polish, and even the older \emph{xlm-roberta-large} model. However, it should be noted that despite weak overall performance, they maintain high quality on tasks involving long texts. This may suggest that changes introduced in these architectures, particularly the application of rotary position embeddings \citep{su2024roformer}, have a positive impact on handling longer contexts. In the small model group, the distilled versions of \emph{polish-roberta-8k} perform well. The best model in this category turned out to be the 12 encoder layers version, which is slightly larger and slower than the other models in the grup. In second place is the version with 6 encoder layers trained progressively, outperforming \emph{polish-roberta-base-v2} in the third place. A detailed comparison of the results between these two models reveals significant differences on two tasks from the KLEJ benchmark: an approximately 9\% worse result on DYK and 4\% on PSC drag down the average scores of the \emph{polish-roberta-8k-6L} model. On the remaining tasks, it achieves results better than or comparable to \emph{polish-roberta-base-v2}. Progressive distillation proved more effective than simple distillation - the directly distilled model performs worse across all task groups except for long texts. As with the larger models, in the smaller group category \emph{EuroBERT-210m} underperforms compared to the other evaluated models.

\vspace{-1em}
\subsection{Bank's internal evaluation}
\vspace{-0.5em}
While the previously described experiments utilized publicly available data, the final phase of this study was conducted using the bank's internal proprietary data. 
The primary objectives were to determine whether: (a) the extended context of the representation model improves classification metrics within the Polish banking domain; and (b) domain adaptation, leveraging several billion tokens from dedicated banking corpora, enhances model quality.

The internal banking corpus comprises approximately 8 billion tokens. This includes roughly 2 billion tokens each of customer email communications, open-source financial texts, and customer complaints with associated responses, alongside approximately 1 billion tokens extracted from OCR-processed banking documents. The remaining data comes from the bank’s public website, intranet resources, expert help-desk documentation, marketing materials, economic journals, and other domain-specific sources.

\begin{table}[h]
  \centering
  \caption{Evaluation on bank's internal classification datasets.}
  \aboverulesep=0ex
  \belowrulesep=0ex
  \setlength{\tabcolsep}{2pt}
  \renewcommand{\arraystretch}{0.8}
  \begin{tabular}{|l|c|c|c|c|}
    \toprule
    \textbf{Model} & \makecell{\textbf{Size} \\ (params)} & \makecell{\textbf{Context} \\ (tokens)} &  \makecell{\textbf{Banking} \\ \textbf{e-mails}}  &  \makecell{\textbf{Mortgage} \\ \textbf{agreements}}\\
    \toprule
    herbert-large-cased (baseline) & 355M & 512 & 82\% & 93\% \\
    polish-roberta-8k (original) & 443M & 8192 & 88\% & 95\% \\
    \rowcolor[HTML]{f0f0f0} polish-roberta-8k (domain-adapted) &  443M & 8192 & \color{blue} 90\% & \color{blue} 95\%\\
    \hline
    \bottomrule
  \end{tabular}
  \label{tab:banking_eval}
\end{table}

The models were fine-tuned on two document classification tasks: banking emails and mortgage loan management documents. The following models were selected for the experiment: \emph{herbert-large-cased}, \emph{polish-roberta-8k}, and the \emph{polish-roberta-8k} model adapted for the banking domain. The results of the experiment are presented in Table \ref{tab:banking_eval}. The email dataset contained 4,675 training and 260 test instances across seven classes: security, complaints, marketing offers, acknowledgments, spam, auto-responses, and miscellaneous messages. The mortgage dataset included 26,059 training and 1,448 test samples distributed among 15 classes of property-related legal documents, including information prospectuses, building permits, zoning plans, cadastral records, bank statements, proofs of funds, Social Insurance Institution certificates, tax authority statements, business activity records, land development decisions, housing cooperative certificates, insurance policies, and Land and Mortgage Register excerpts. Results indicate that the domain-adapted \emph{polish-roberta-8k} model outperforms both the original model and the baseline HerBERT model. Specifically, on the Banking Emails Dataset, an improvement of approximately 8 percentage points over the benchmark was observed. Regarding mortgage document classification, the domain-adapted model performed comparably to the standard \emph{polish-roberta-8k}, yet significantly surpassed the HerBERT baseline. In summary, both domain adaptation and an extended context window contribute to enhanced classification performance within the Polish banking domain.

\vspace{-1em}
\section{Conclusions}
\vspace{-0.5em}
In this publication, we introduced \emph{polish-roberta-8k}, a Polish-language encoder model with an extended context length, enabling processing of long documents efficiently without compromising prediction quality. The training strategy, including two-stage weight updates and selective checkpoint merging, proved effective in balancing performance across both short and long contexts. We also explored model compression via knowledge distillation, producing smaller variants with 12 and 6 encoder layers. Progressive distillation was particularly effective in preserving model quality in the smallest variant. Evaluation across 25 diverse tasks, including the KLEJ benchmark, the newly introduced financial benchmark FinBench, and other Polish-language datasets, demonstrates that \emph{polish-roberta-8k} achieves state-of-the-art performance among large models and outperforms comparable models on long-text tasks.

\vspace{1mm}
\noindent \textbf{Acknowledgments}: This project is financed by the European Funds, registered under the number FENG.01.01-IP.01-A028/23-00. It focuses on "Building innovative large language models and a service platform for serving multi-task models within the Bank". We would like to give special thanks to the members of PKO team: Olga Walenciuk, Łukasz Wójcik, Szymon Balawajder, Marcin Kowalski, Tomasz Groszkowski, Grzegorz Piekarzewski.

\bibliographystyle{splncsnat}
\bibliography{references}

@misc{ociepa2025bielik11bv2technical,
      title={Bielik 11B v2 Technical Report}, 
      author={Krzysztof Ociepa and Łukasz Flis and Krzysztof Wróbel and Adrian Gwoździej and Remigiusz Kinas},
      year={2025},
      eprint={2505.02410},
      archivePrefix={arXiv},
      primaryClass={cs.CL},
      url={https://arxiv.org/abs/2505.02410}, 
}

@article{kocon2025pllum,
  title={PLLuM: A Family of Polish Large Language Models},
  author={Koco{\'n}, Jan and Piasecki, Maciej and Janz, Arkadiusz and Ferdinan, Teddy and Radli{\'n}ski, {\L}ukasz and Koptyra, Bart{\l}omiej and Oleksy, Marcin and Wo{\'z}niak, Stanis{\l}aw and Walkowiak, Pawe{\l} and Wojtasik, Konrad and others},
  journal={arXiv preprint arXiv:2511.03823},
  year={2025}
}

@Misc{polish-nlp-resources,
  author =       {S{\l}awomir Dadas},
  title =        {A repository of Polish {NLP} resources},
  howpublished = {Github},
  year =         {2019},
  url =          {https://github.com/sdadas/polish-nlp-resources/}
}

@inproceedings{szmyd-etal-2023-trelbert,
    title = "{T}rel{BERT}: A pre-trained encoder for {P}olish {T}witter",
    author = "Szmyd, Wojciech  and
      Kotyla, Alicja  and
      Zobni{\'o}w, Micha{\l}  and
      Falkiewicz, Piotr  and
      Bartczuk, Jakub  and
      Zygad{\l}o, Artur",
    booktitle = "Proceedings of the 9th Workshop on Slavic Natural Language Processing 2023 (SlavicNLP 2023)",
    month = may,
    year = "2023",
    publisher = "Association for Computational Linguistics",
    doi = "10.18653/v1/2023.bsnlp-1.3",
    pages = "17--24"
}

@inproceedings{chrabrowa2022evaluation,
  title={Evaluation of Transfer Learning for Polish with a Text-to-Text Model},
  author={Chrabrowa, Aleksandra and Dragan, {\L}ukasz and Grzegorczyk, Karol and Kajtoch, Dariusz and Koszowski, Miko{\l}aj and Mroczkowski, Robert and Rybak, Piotr},
  booktitle={Proceedings of the Thirteenth Language Resources and Evaluation Conference},
  year={2022}
}

@article{portes2023mosaicbert,
  title={MosaicBERT: A bidirectional encoder optimized for fast pretraining},
  author={Portes, Jacob and Trott, Alexander and Havens, Sam and King, Daniel and Venigalla, Abhinav and Nadeem, Moin and Sardana, Nikhil and Khudia, Daya and Frankle, Jonathan},
  journal={Advances in Neural Information Processing Systems},
  volume={36},
  pages={3106--3130},
  year={2023}
}

@article{vaswani2017attention,
  title={Attention is all you need},
  author={Vaswani, Ashish and Shazeer, Noam and Parmar, Niki and Uszkoreit, Jakob and Jones, Llion and Gomez, Aidan N and Kaiser, {\L}ukasz and Polosukhin, Illia},
  journal={Advances in neural information processing systems},
  volume={30},
  year={2017}
}

@article{kocon2023chatgpt,
  title={ChatGPT: Jack of all trades, master of none},
  author={Koco{\'n}, Jan and Cichecki, Igor and Kaszyca, Oliwier and Kochanek, Mateusz and Szyd{\l}o, Dominika and Baran, Joanna and Bielaniewicz, Julita and Gruza, Marcin and Janz, Arkadiusz and Kanclerz, Kamil and others},
  journal={Information fusion},
  volume={99},
  pages={101861},
  year={2023},
  publisher={Elsevier}
}

@article{brown2020language,
  title={Language models are few-shot learners},
  author={Brown, Tom and Mann, Benjamin and Ryder, Nick and Subbiah, Melanie and Kaplan, Jared D and Dhariwal, Prafulla and Neelakantan, Arvind and Shyam, Pranav and Sastry, Girish and Askell, Amanda and others},
  journal={Advances in neural information processing systems},
  volume={33},
  pages={1877--1901},
  year={2020}
}

@article{raiaan2024review,
  title={A review on large language models: Architectures, applications, taxonomies, open issues and challenges},
  author={Raiaan, Mohaimenul Azam Khan and Mukta, Md Saddam Hossain and Fatema, Kaniz and Fahad, Nur Mohammad and Sakib, Sadman and Mim, Most Marufatul Jannat and Ahmad, Jubaer and Ali, Mohammed Eunus and Azam, Sami},
  journal={IEEE access},
  volume={12},
  pages={26839--26874},
  year={2024},
  publisher={IEEE}
}

@article{raffel2020exploring,
  title={Exploring the limits of transfer learning with a unified text-to-text transformer},
  author={Raffel, Colin and Shazeer, Noam and Roberts, Adam and Lee, Katherine and Narang, Sharan and Matena, Michael and Zhou, Yanqi and Li, Wei and Liu, Peter J},
  journal={Journal of machine learning research},
  volume={21},
  number={140},
  pages={1--67},
  year={2020}
}

@article{su2024roformer,
  title={Roformer: Enhanced transformer with rotary position embedding},
  author={Su, Jianlin and Ahmed, Murtadha and Lu, Yu and Pan, Shengfeng and Bo, Wen and Liu, Yunfeng},
  journal={Neurocomputing},
  volume={568},
  year={2024},
  publisher={Elsevier}
}

@inproceedings{warner2025smarter,
  title={Smarter, better, faster, longer: A modern bidirectional encoder for fast, memory efficient, and long context finetuning and inference},
  author={Warner, Benjamin and Chaffin, Antoine and Clavi{\'e}, Benjamin and Weller, Orion and Hallstr{\"o}m, Oskar and Taghadouini, Said and Gallagher, Alexis and Biswas, Raja and Ladhak, Faisal and Aarsen, Tom and others},
  booktitle={Proceedings of the 63rd Annual Meeting of the Association for Computational Linguistics (Volume 1: Long Papers)},
  pages={2526--2547},
  year={2025}
}

@misc{marone2025mmbertmodernmultilingualencoder,
      title={mmBERT: A Modern Multilingual Encoder with Annealed Language Learning}, 
      author={Marc Marone and Orion Weller and William Fleshman and Eugene Yang and Dawn Lawrie and Benjamin Van Durme},
      year={2025},
      eprint={2509.06888},
      archivePrefix={arXiv},
      primaryClass={cs.CL},
      url={https://arxiv.org/abs/2509.06888}, 
}

@inproceedings{boizard2025eurobert,
title={Euro{BERT}: Scaling Multilingual Encoders for European Languages},
author={Nicolas Boizard and Hippolyte Gisserot-Boukhlef and Duarte Miguel Alves and Andre Martins and Ayoub Hammal and Caio Corro and CELINE HUDELOT and Emmanuel Malherbe and Etienne Malaboeuf and Fanny Jourdan and Gabriel Hautreux and Jo{\~a}o Alves and Kevin El Haddad and Manuel Faysse and Maxime Peyrard and Nuno M Guerreiro and Patrick Fernandes and Ricardo Rei and Pierre Colombo},
booktitle={Second Conference on Language Modeling},
year={2025},
url={https://openreview.net/forum?id=jdOC24msVq}
}

@inproceedings{conneau2020unsupervised,
  title={Unsupervised cross-lingual representation learning at scale},
  author={Conneau, Alexis and Khandelwal, Kartikay and Goyal, Naman and Chaudhary, Vishrav and Wenzek, Guillaume and Guzm{\'a}n, Francisco and Grave, Edouard and Ott, Myle and Zettlemoyer, Luke and Stoyanov, Veselin},
  booktitle={Proceedings of the 58th annual meeting of the association for computational linguistics},
  year={2020}
}

@inproceedings{mroczkowski-etal-2021-herbert,
    title = "{H}er{BERT}: Efficiently Pretrained Transformer-based Language Model for {P}olish",
    author = "Mroczkowski, Robert  and
      Rybak, Piotr  and
      Wr{\'o}blewska, Alina  and
      Gawlik, Ireneusz",
    editor = "Babych, Bogdan  and
      Kanishcheva, Olga  and
      Nakov, Preslav  and
      Piskorski, Jakub  and
      Pivovarova, Lidia  and
      Starko, Vasyl  and
      Steinberger, Josef  and
      Yangarber, Roman  and
      Marci{\'n}czuk, Micha{\l}  and
      Pollak, Senja  and
      P{\v{r}}ib{\'a}{\v{n}}, Pavel  and
      Robnik-{\v{S}}ikonja, Marko",
    booktitle = "Proceedings of the 8th Workshop on Balto-Slavic Natural Language Processing",
    month = apr,
    year = "2021",
    address = "Kiyv, Ukraine",
    publisher = "Association for Computational Linguistics",
    url = "https://aclanthology.org/2021.bsnlp-1.1/",
    pages = "1--10"
}

@inproceedings{sinha2021impact,
  title={Impact of news on the commodity market: Dataset and results},
  author={Sinha, Ankur and Khandait, Tanmay},
  booktitle={Future of Information and Communication Conference},
  pages={589--601},
  year={2021},
  organization={Springer}
}

@article{malo2014good,
  title={Good debt or bad debt: Detecting semantic orientations in economic texts},
  author={Malo, Pekka and Sinha, Ankur and Korhonen, Pekka and Wallenius, Jyrki and Takala, Pyry},
  journal={Journal of the Association for Information Science and Technology},
  volume={65},
  number={4},
  pages={782--796},
  year={2014},
  publisher={Wiley Online Library}
}

@inproceedings{maia201818,
  title={Www'18 open challenge: financial opinion mining and question answering},
  author={Maia, Macedo and Handschuh, Siegfried and Freitas, Andr{\'e} and Davis, Brian and McDermott, Ross and Zarrouk, Manel and Balahur, Alexandra},
  booktitle={Companion proceedings of the the web conference 2018},
  year={2018}
}

@inproceedings{maas2011learning,
  title={Learning word vectors for sentiment analysis},
  author={Maas, Andrew and Daly, Raymond E and Pham, Peter T and Huang, Dan and Ng, Andrew Y and Potts, Christopher},
  booktitle={Proceedings of the 49th annual meeting of the association for computational linguistics: Human language technologies},
  pages={142--150},
  year={2011}
}

@inproceedings{kolos2024ban,
  title={BAN-PL: A Polish dataset of banned harmful and offensive content from wykop. pl web service},
  author={Ko{\l}os, Anna and Okulska, Inez and G{\l}{\k{a}}bi{\'n}ska, Kinga and Karli{\'n}ska, Agnieszka and Wi{\'s}nios, Emilia and Ellerik, Pawe{\l} and Pra{\l}at, Andrzej},
  booktitle={Proceedings of the 2024 Joint International Conference on Computational Linguistics, Language Resources and Evaluation (LREC-COLING 2024)},
  pages={2107--2118},
  year={2024}
}

@inproceedings{modzelewski2024mipd,
  title={Mipd: Exploring manipulation and intention in a novel corpus of polish disinformation},
  author={Modzelewski, Arkadiusz and Da San Martino, Giovanni and Savov, Pavel and Wilczy{\'n}ska, Magdalena Anna and Wierzbicki, Adam},
  booktitle={Proceedings of the 2024 Conference on Empirical Methods in Natural Language Processing},
  pages={19769--19785},
  year={2024}
}

@inproceedings{bogdanowicz2023twitteremo,
  title={Twitteremo: Annotating emotions and sentiment in polish twitter},
  author={Bogdanowicz, Stanis{\l}aw and Cwynar, Hanna and Zwierzchowska, Aleksandra and Klamra, Cezary and Kiera{\'s}, Witold and Kobyli{\'n}ski, {\L}ukasz},
  booktitle={International Conference on Computational Science},
  pages={212--220},
  year={2023},
  organization={Springer}
}

@inproceedings{chalkidis-etal-2019-large,
    title = "Large-Scale Multi-Label Text Classification on {EU} Legislation",
    author = "Chalkidis, Ilias  and Fergadiotis, Manos  and Malakasiotis, Prodromos  and Androutsopoulos, Ion",
    booktitle = "Proceedings of the 57th Annual Meeting of the Association for Computational Linguistics",
    year = "2019",
    address = "Florence, Italy",
    publisher = "Association for Computational Linguistics",
    url = "https://www.aclweb.org/anthology/P19-1636",
    doi = "10.18653/v1/P19-1636",
    pages = "6314--6322"
}

@inproceedings{dadas2022training,
  title={Training effective neural sentence encoders from automatically mined paraphrases},
  author={Dadas, S{\l}awomir},
  booktitle={2022 IEEE International Conference on Systems, Man, and Cybernetics (SMC)},
  pages={371--378},
  year={2022},
  organization={IEEE}
}

@inproceedings{dadas2020evaluation,
  title={Evaluation of sentence representations in Polish},
  author={Dadas, Slawomir and Pere{\l}kiewicz, Micha{\l} and Po{\'s}wiata, Rafa{\l}},
  booktitle={Proceedings of the Twelfth Language Resources and Evaluation Conference},
  pages={1674--1680},
  year={2020}
}

@article{liu2024deepseek,
  title={Deepseek-v3 technical report},
  author={Liu, Aixin and Feng, Bei and Xue, Bing and Wang, Bingxuan and Wu, Bochao and Lu, Chengda and Zhao, Chenggang and Deng, Chengqi and Zhang, Chenyu and Ruan, Chong and others},
  journal={arXiv preprint arXiv:2412.19437},
  year={2024}
}

@inproceedings{dadas2020pre,
  title={Pre-training polish transformer-based language models at scale},
  author={Dadas, S{\l}awomir and Pere{\l}kiewicz, Micha{\l} and Po{\'s}wiata, Rafa{\l}},
  booktitle={International Conference on Artificial Intelligence and Soft Computing},
  pages={301--314},
  year={2020},
  organization={Springer}
}

@inproceedings{rybak2020klej,
  title={KLEJ: Comprehensive Benchmark for Polish Language Understanding},
  author={Rybak, Piotr and Mroczkowski, Robert and Tracz, Janusz and Gawlik, Ireneusz},
  booktitle={Proceedings of the 58th Annual Meeting of the Association for Computational Linguistics},
  year={2020}
}

@inproceedings{devlin2019bert,
  title={Bert: Pre-training of deep bidirectional transformers for language understanding},
  author={Devlin, Jacob and Chang, Ming-Wei and Lee, Kenton and Toutanova, Kristina},
  booktitle={Proceedings of the 2019 conference of the North American chapter of the association for computational linguistics: human language technologies, volume 1 (long and short papers)},
  pages={4171--4186},
  year={2019}
}

@article{liu2019roberta,
  title={Roberta: A robustly optimized bert pretraining approach},
  author={Liu, Yinhan and Ott, Myle and Goyal, Naman and Du, Jingfei and Joshi, Mandar and Chen, Danqi and Levy, Omer and Lewis, Mike and Zettlemoyer, Luke and Stoyanov, Veselin},
  journal={arXiv preprint arXiv:1907.11692},
  year={2019}
}

@article{breton2025neobert,
  title={NeoBERT: A Next-Generation BERT},
  author={Breton, Lola Le and Fournier, Quentin and Mezouar, Mariam El and Morris, John X and Chandar, Sarath},
  journal={arXiv preprint arXiv:2502.19587},
  year={2025}
}

@article{dao2022flashattention,
  title={Flashattention: Fast and memory-efficient exact attention with io-awareness},
  author={Dao, Tri and Fu, Dan and Ermon, Stefano and Rudra, Atri and R{\'e}, Christopher},
  journal={Advances in neural information processing systems},
  volume={35},
  pages={16344--16359},
  year={2022}
}

@inproceedings{wang2023distill,
  title={How to Distill your BERT: An Empirical Study on the Impact of Weight Initialisation and Distillation Objectives},
  author={Wang, Xinpeng and Weissweiler, Leonie and Sch{\"u}tze, Hinrich and Plank, Barbara},
  booktitle={Proceedings of the 61st Annual Meeting of the Association for Computational Linguistics (Volume 2: Short Papers)},
  pages={1843--1852},
  year={2023}
}

@article{wang2020minilm,
  title={Minilm: Deep self-attention distillation for task-agnostic compression of pre-trained transformers},
  author={Wang, Wenhui and Wei, Furu and Dong, Li and Bao, Hangbo and Yang, Nan and Zhou, Ming},
  journal={Advances in neural information processing systems},
  volume={33},
  year={2020}
}

@inproceedings{Casanueva2020,
    author      = {I{\~{n}}igo Casanueva and Tadas Temcinas and Daniela Gerz and Matthew Henderson and Ivan Vulic},
    title       = {Efficient Intent Detection with Dual Sentence Encoders},
    year        = {2020},
    month       = {mar},
    note        = {Data available at https://github.com/PolyAI-LDN/task-specific-datasets},
    url         = {https://arxiv.org/abs/2003.04807},
    booktitle   = {Proceedings of the 2nd Workshop on NLP for ConvAI - ACL 2020}
}

\end{document}